\pdfoutput=1

\documentclass[11pt]{article}

\usepackage{emnlp2021}

\usepackage{times}
\usepackage{latexsym}

\usepackage[T1]{fontenc}

\usepackage[utf8]{inputenc}

\usepackage{microtype}

\usepackage{amsmath}
\usepackage{amsfonts}
\usepackage{booktabs}
\usepackage{multicol}
\usepackage{multirow}
\usepackage{graphicx}
\usepackage{subcaption}

\newcommand\ti[1]{\textit{#1}}

\newcommand\tf[1]{\textbf{#1}}

\usepackage{amsmath,amsfonts,bm}

\def\1{\bm{1}}

\makeatletter
\newcommand{\ve}{\@ifnextchar\bgroup{\velong}{{\bm{e}}}}
\newcommand{\velong}[1]{{\bm{#1}}}
\makeatother

\def\ve{{\mathbf{e}}}

\DeclareMathAlphabet{\mathsfit}{\encodingdefault}{\sfdefault}{m}{sl}
\SetMathAlphabet{\mathsfit}{bold}{\encodingdefault}{\sfdefault}{bx}{n}

\def\gD{{\mathcal{D}}}

\def\gS{{\mathcal{S}}}
\def\gT{{\mathcal{T}}}

\def\gV{{\mathcal{V}}}

\newcommand{\E}{\mathbb{E}}

\DeclareMathOperator*{\argmin}{arg\,min}

\newcommand{\squad}{SQuAD 1.1}

\newcommand{\roberta}{Single-dataset fine-tuning}

\newcommand{\unifiedqa}{UnifiedQA-base}
\newcommand{\robertamulti}{Multi-dataset fine-tuning}
\newcommand{\adapter}{Single-dataset adapters}
\newcommand{\multiadapter}{MADE}
\newcommand{\preaverage}{pre avg.}
\newcommand{\postaverage}{post avg.}

\title{Single-dataset Experts for Multi-dataset Question Answering}

\author{Dan Friedman \quad Ben Dodge \quad Danqi Chen \\\textit{}
Department of Computer Science, Princeton, NJ \\
\texttt{\{dfriedman,bd4,danqic\}@cs.princeton.edu}}

\begin{document}
\maketitle

\begin{abstract}
Many datasets have been created for training reading comprehension models, and a natural question is whether we can combine them to build models that (1) perform better on all of the training datasets and (2) generalize and transfer better to new datasets.
Prior work has addressed this goal by training one network simultaneously on multiple datasets, which works well on average but is prone to over- or under-fitting different sub-distributions and might transfer worse compared to source models with more overlap with the target dataset.
Our approach is to model multi-dataset question answering with a collection of single-dataset experts, by training a collection of lightweight, dataset-specific adapter modules~\citep{houlsby2019parameter} that share an underlying Transformer model.
We find that these \tf{M}ulti-\tf{A}dapter \tf{D}ataset \tf{E}xperts ({\multiadapter}) outperform all our baselines in terms of in-distribution accuracy, and simple methods based on parameter-averaging lead to better zero-shot generalization and few-shot transfer performance,
offering a strong and versatile starting point for building new reading comprehension systems.\footnote{Our code and models are available at \url{https://github.com/princeton-nlp/MADE}.
}

\end{abstract}

\begin{figure}[ht!]
    \centering
    \resizebox{0.8\columnwidth}{!}{%
    \includegraphics{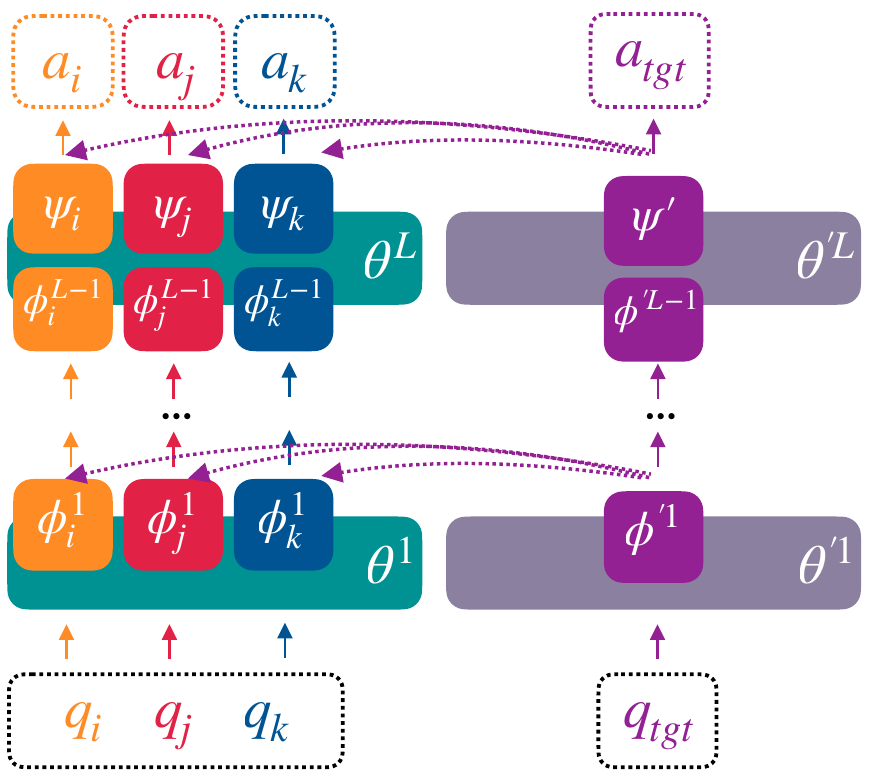}
    }
    \caption{
    MADE consists of a set of dataset-specific adapters and classifiers $\{(\phi_t, \psi_t)\}_{t=1}^{|\gS|}$ with a shared Transformer model $\theta$.
    They are optimized jointly on a set of training datasets (left). To transfer to a new dataset (right), we either average the parameters of the adapters, or fine-tune all adapters on the target dataset and take the weighted average at the end (Section~\ref{sec:method}).
    }
    \label{fig:multiadapter_figure}
\end{figure}
\vspace{-1em}

\section{Introduction}

The goal of reading comprehension is to create computer programs that can answer questions based on a single passage of text.
Many reading comprehension datasets have been introduced over the years, and prior work has explored ways of training one network on multiple datasets to get a model that generalizes better to new distributions~\citep{talmor2019multiqa,fisch2019mrqa,khashabi2020unifiedqa}.
Our goal is to build a multi-dataset model that performs well on the training distributions and can also serve as a strong starting point for transfer learning to new datasets.
Multi-dataset training provides a way to model the regularities between datasets but it has the following shortcomings.
First, multi-task models are liable to over- or under-fit different tasks~\citep{gottumukkala2020dynamic}, which can result in worse in-distribution accuracy.
Second, given a particular target dataset, multi-dataset models might achieve worse transfer performance compared to a specialized model trained on a more similar source dataset.

Our idea is to combine the benefits of single- and multi-dataset by training a collection of single-dataset experts that share an underlying Transformer model~(Figure~\ref{fig:multiadapter_figure}).
This system is based on adapters~\citep{houlsby2019parameter}, lightweight task-specific modules interleaved between the layers of a pre-trained Transformer (e.g., BERT;~\citealp{devlin2018bert}).
The standard use of adapters is as a parameter-efficient alternative to fine-tuning: task-specific adapters are trained separately on top of a frozen Transformer, which means the adapters cannot directly learn cross-task regularities.
We instead first train a shared Transformer in a multi-adapter setup before refining adapters for individual datasets, which we call \tf{M}ulti-\tf{A}dapter \tf{D}ataset \tf{E}xperts ({\multiadapter}). Our intuition is that the shared parameters encode regularities across different reading comprehension tasks while the adapters model the sub-distributions, resulting in more accurate and robust specialized models that transfer better to a variety of target datasets.

We apply this approach to a range of extractive question answering datasets from the MRQA 2019 shared task~\citep{fisch2019mrqa}, training {\multiadapter} on six in-domain datasets and evaluating generalization and few-shot transfer learning to six out-of-domain datasets.
The resulting system outperforms single- and multi-dataset models in terms of in-domain accuracy, and we find that a simple approach to transfer learning works well: averaging the parameters of the {\multiadapter} adapters results in a single model that gets better zero-shot generalization and few-shot transfer performance compared to both baselines as well as a state-of-the-art multi-dataset QA model, UnifiedQA~\citep{khashabi2020unifiedqa}.
Our experiments illustrate the benefits of modeling both cross-dataset regularities and dataset-specific attributes, and the trained models offer a strong and versatile starting point for new question-answering models.

\section{Related Work}

Prior work has addressed multi-dataset reading comprehension by fine-tuning a pre-trained Transformer language model~\citep{devlin2018bert} simultaneously on examples from multiple datasets~\citep{talmor2019multiqa,fisch2019mrqa}.
Several works explore different multi-task sampling schedules, as a way of mitigating training set imbalances~\citep{xu2019multi,gottumukkala2020dynamic}.
Another line of work focuses on training models to answer a wider variety of question types, including UnifiedQA~\citep{khashabi2020unifiedqa}, a T5 model~\cite{raffel2020exploring} trained on datasets with different answer formats, such as yes/no and multiple-choice, using a unified text-to-text format.

Adapters~\citep{houlsby2019parameter,rebuffi2018efficient} are task-specific modules interleaved between the layers of a shared Transformer.%
~\citet{stickland2019bert} trained task adapters and the Transformer parameters jointly for the GLUE benchmark~\citep{wang2019glue} but achieved mixed results, improving on small datasets but degrading on larger ones.
Subsequent work has used a frozen, pre-trained Transformer and trained task adapters separately.
Researchers have explored different methods for achieving transfer learning in this setting, such as learning to interpolate the activations of pre-trained adapters~\citep{pfeiffer2021adapterfusion}.

\begin{table*}[ht!]
\centering
\resizebox{\linewidth}{!}{%
\begin{tabular}{l | ccccccc }
\toprule
\tf{Model} & \tf{\squad} & \tf{HotpotQA} & \tf{TriviaQA} & \tf{NewsQA} & \tf{SearchQA} & \tf{NaturalQ} & \tf{Avg.}\\
\midrule
Dynamic sampling & 91.4 & 80.9 & \tf{80.5} & 71.4 & 84.6 & 79.8 & 81.4 \\
\midrule
\roberta & 90.9 & 78.6 & 79.3 & 70.3 & 84.5 & 79.2 & 80.5 \\
\adapter & 91.4 & 78.5 & 79.6 & 70.9 & 85.1 & 79.2 & 80.8 \\
\robertamulti & 91.8 & 81.0 & 80.1 & \tf{72.3} & 84.7 & 79.5 & 81.6 \\
{\multiadapter} (w/o adapter tuning) & 91.9 & 80.7 & 80.1 & 71.8 & 84.5 & 79.5 & 81.4 \\
{\multiadapter} (w/ adapter tuning) & \tf{92.4} & \tf{81.5} & \tf{80.5} & 72.1 & \tf{85.8} & \tf{80.9} & \tf{82.2} \\
\bottomrule
\end{tabular}
}
\caption{
\label{tab:in_domain}
In-domain F1.
\ti{Dynamic sampling} is a reimplementation of the method from~\citet{gottumukkala2020dynamic} on the MRQA datasets.
We compare {\multiadapter} results at the end of joint optimization (\ti{w/o adapter tuning}), and after freezing the Transformer and tuning the adapters separately (\ti{w/ adapter tuning}).
See Section~\ref{sec:in_distribution} for details.
}
\end{table*}

\section{Method}
\label{sec:method}

\subsection{Problem Definition}
\label{sec:problem_definition}
The objective of reading comprehension is to model the distribution $p(a \mid q, c)$, where $q, c, a \in \gV^*$ represent a \ti{question}, supporting \ti{context}, and \ti{answer} respectively and consist of sequences of tokens from a vocabulary $\gV$.
For simplicity, we focus on extractive reading comprehension, where every question can be answered by selecting a span of tokens in the context, but the approach is generic and can be extended to other formats.
We make the standard assumption that the probability of context span $c_{i \ldots j}$ being the answer can be decomposed into the product of $p(\text{start} = i \mid q, c)$ and $p(\text{end} = j \mid q, c)$.

We consider a collection of source datasets $\gS$ and target datasets $\gT$, where each dataset $\gD \in \gS \cup \gT$ consists of supervised examples in the form $(q, c, a)$.
The goal is to train a model on $\gS$ that achieves high in-domain accuracy and transfers well to unseen datasets in $\gT$, either zero-shot or given a small number of labeled examples.

\subsection{Multi-dataset Fine-tuning}
The standard approach to multi-dataset reading comprehension is to fit a single model to examples drawn uniformly from the datasets in $\gS$:
\[
  \argmin_{\theta, \psi} \E_{\gD_i \sim \gS} \left [
   \E_{q, c, a \sim \gD_i}[ - \log p_{\theta, \psi}(a \mid q, c)]
 \right ],
\]
where $\theta$ refers to the parameters of an encoder model (usually a pre-trained Transformer like BERT; \citealp{devlin2018bert}), which maps a question and context to a sequence of contextualized token embeddings, and $\psi$ denotes the classifier weights used to predict the start and end tokens.

The objective is approximated by training on mixed mini-batches with approximately equal numbers of examples from each dataset~\citep{fisch2019mrqa,khashabi2020unifiedqa},
although some researchers have investigated more sophisticated sampling strategies~\citep{xu2019multi}.
For example,~\citet{gottumukkala2020dynamic} introduce \ti{dynamic sampling}, sampling from each dataset in inverse proportion to the model's current validation accuracy.

\subsection{\multiadapter}

Our approach is to explicitly model the fact that our data represent a mixture of datasets.
We decompose the model parameters into a shared Transformer, $\theta$,
and dataset-specific token classifiers $\boldsymbol{\psi} = \psi_1, \ldots, \psi_{|\gS|}$ and adapters $\boldsymbol{\phi} = \phi_1, \ldots, \phi_{|\gS|}$ (Figure~\ref{fig:multiadapter_figure}).
We use a two-phase optimization procedure to fit these parameters.
In the \ti{joint optimization} phase, we jointly train all of the parameters on the source datasets:
\begin{align*}
&\argmin_{\theta, \boldsymbol{\phi}, \boldsymbol{\psi}} \E_{\gD_i \sim \gS} \left [
\E_{q, c, a \sim \gD_i}[- \log p_{\theta, \phi_i, \psi_i}(a \mid q, c)]
\right ]
\end{align*}
After validation accuracy (average F1 scores of the source datasets) stops improving,
we freeze $\theta$ and continue \ti{adapter tuning}, refining each pair of $(\phi_i, \psi_i)$  separately on each dataset.

\paragraph{Zero-shot generalization}
We use a simple strategy to extend {\multiadapter} to an unseen dataset: we initialize a new adapter and classifier $(\phi', \psi')$ by averaging the parameters of the pre-trained adapters and classifiers $\phi_1, \ldots, \phi_{|\gS|}$ and $\psi_1, \ldots, \psi_{|\gS|}$, and return the answer with the highest probability under $p_{\theta, \phi', \psi'}(a \mid q, c)$.
We also considered an ensemble approach, averaging the token-level probabilities predicted by each adapter, but found this to perform similarly to parameter averaging at the additional cost of running the model $|\gS|$ times.

\paragraph{Transfer learning}
We also consider a transfer learning setting, in which a small number of labeled examples of a target domain (denoted $\gD_{\text{tgt}}$) are provided. We explore two ways to build a single, more accurate model.
The first is to initialize $(\phi', \psi')$ as a weighted average of pre-trained adapters, $\phi' = \frac{1}{|\gS|} \sum_{i=1}^{|\gS|} \alpha_i \phi_i$, and classifiers $\psi' = \frac{1}{|\gS|} \sum_{i=1}^{|\gS|} \alpha_i \psi_i$, using $\gD_{\text{tgt}}$ to estimate the mixture weights.
For each $i$, we set the mixture weight $\alpha_i$ to be proportional to the exponential of the negative zero-shot loss on the training data:
\[
\alpha_i \propto \mathrm{exp} \left (
\E_{q, c, a \in \gD_{\text{tgt}}}[\log p_{\theta, \phi_i, \psi_i}(a \mid q, c)]
\right ),
\]
and then tune $\theta$ and $(\phi', \psi')$ on the target dataset.
The second approach is to first jointly tune $\theta$, $\boldsymbol{\phi}$, and $\boldsymbol{\psi}$ on $\gD_{\text{tgt}}$, maximizing the marginal likelihood: \[
\E_{q, c, a \sim \gD_{\text{tgt}}} \left [\log \frac{1}{|\gS|} \sum_{i=1}^{|\gS|} p_{\theta, \phi_i, \psi_i}(a \mid q, c) \right ],
\]
and then take the weighted average of the parameters, calculating the mixture weights $\alpha_i$ as above but using the loss of the fine-tuned adapters on a small number of held-out examples from $\gD_{\text{tgt}}$.
Pre-averaging is faster to train, because it only involves training one model rather than all $|\gS|$ adapters.
After training, both approaches result in a single model that only requires running one forward pass through $(\theta, \phi', \psi')$ to make a prediction.

\begin{table*}[t!]
\centering
\resizebox{\linewidth}{!}{%
\begin{tabular}{l | ccccccc }
\toprule
\tf{Model} & \tf{BioASQ} & \tf{DROP} & \tf{DuoRC} & \tf{RACE} & \tf{RelEx} & \tf{TextbookQA} & \tf{Avg.} \\
 \midrule
\unifiedqa & 59.7 & 45.7 & 30.4 & 51.4$^\dagger$ & 82.0 & 35.9 & 50.9 \\
\midrule
\robertamulti & 64.1 & 51.5 & 63.0 & 47.6 & 87.3 & 59.0 & 62.1 \\
{\adapter} & 53.4 & 20.5 & 49.4 & 22.3 & 72.7 & 37.8 & 42.7 \\
{\multiadapter} (w/o adapter-tuning) & 66.5 & 50.9 & \tf{67.2} & 47.8 & 86.7 & 58.5 & 62.9 \\
{\multiadapter} (w/ adapter-tuning) & \tf{66.6} & \tf{52.2} & 66.9 & \tf{48.0} & \tf{87.6} & \tf{60.0} & \tf{63.5} \\
\bottomrule
\end{tabular}
}
\caption{
\label{tab:zero_shot}
Zero-shot generalization results (F1), averaging parameters of the dataset-specific adapters, and comparing {\multiadapter} at the end of joint optimization (\ti{w/o adapter tuning}), and after tuning the adapters separately (\ti{w/ adapter tuning}). See Section~\ref{sec:zero_shot}.
$^\dagger$: RACE is part of the UnifiedQA training data.
}
\end{table*}

\begin{figure*}
    \centering
    \resizebox{0.9\linewidth}{!}{%
    \includegraphics{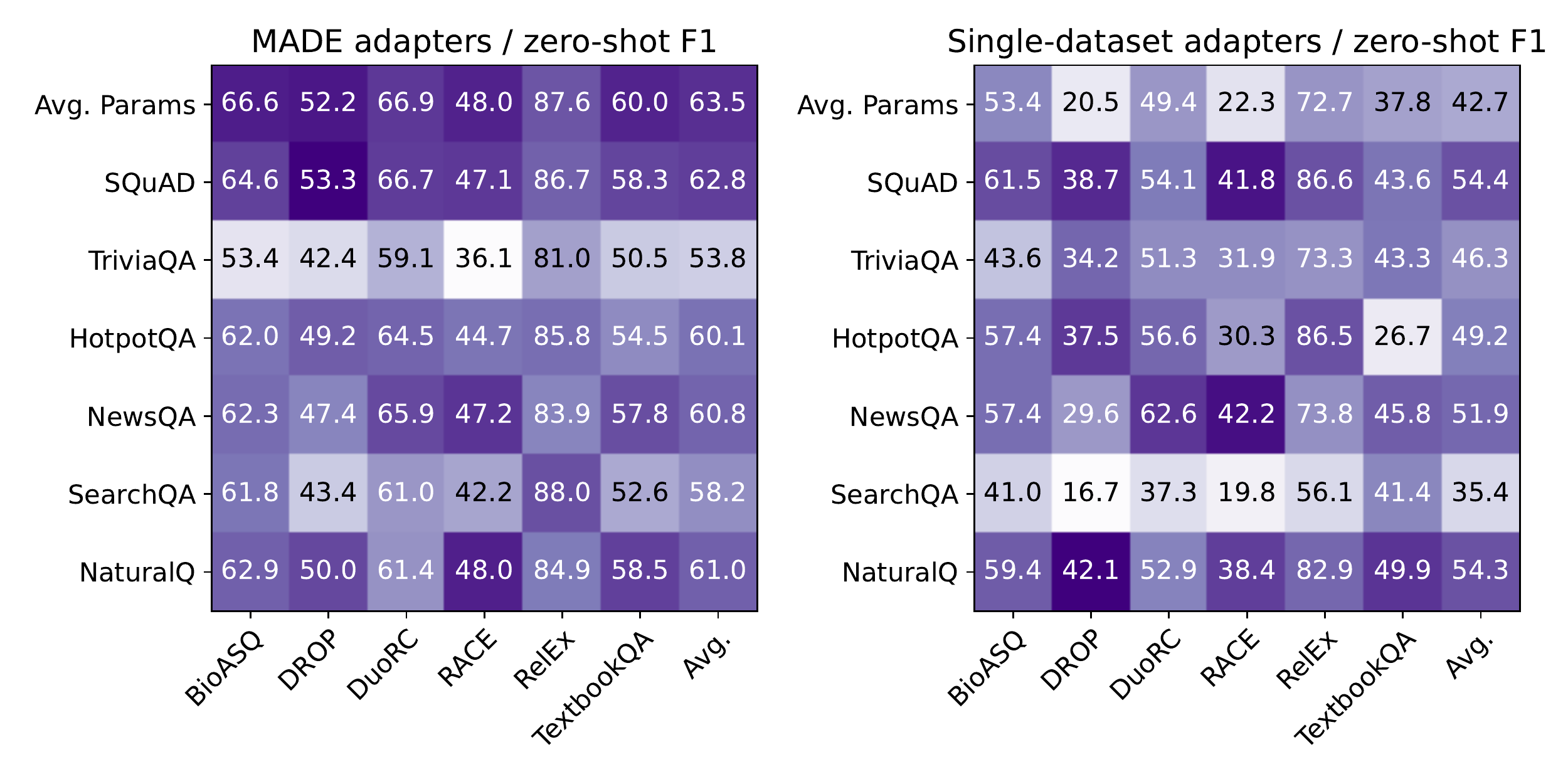}
    }
    \caption{
    Out-of-domain accuracy (F1) of the different MADE adapters (left) compared with single-dataset adapters (right), which are trained trained separately with a frozen Transformer.
    \ti{Avg. Params:} An adapter created by averaging the parameters of the dataset-specific adapters.
    }
    \label{fig:zero_shot_comparison}
\end{figure*}

\section{Experiments}
\label{sec:experiments}
\subsection{Setup}
We use the datasets from the MRQA 2019 shared task~\citep{fisch2019mrqa}, which are split into six large in-domain datasets,\footnote{
SQuAD 1.1~\citep{rajpurkar2016squad}, HotpotQA~\citep{yang2018hotpotqa}, TriviaQA~\citep{joshi2017triviaqa}, NewsQA~\citep{trischler2017newsqa}, SearchQA~\citep{dunn2017searchqa}, and Natural Questions~\citep{kwiatkowski2019natural}
} and six small out-of-domain datasets.\footnote{
BioASQ~\citep{tsatsaronis2015overview}, DROP~\citep{dua2019drop}, DuoRC~\citep{saha2018duorc}, RACE~\citep{lai2017race}, RelationExtraction~\citep{levy2017zero}, and TextbookQA~\citep{kembhavi2017you}.
}
Dataset statistics are in Appendix~\ref{appendix:dataset_details}.
We use the RoBERTa-base model~\citep{liu2019roberta} with the default adapter configuration from~\citet{houlsby2019parameter}, which adds approximately 1.8M parameters to the {\textasciitilde}128M in RoBERTa-base (1\%).

\subsection{In-domain Performance}

\label{sec:in_distribution}
First we train {\multiadapter} on the six training datasets and compare in-domain accuracy with single- and multi-dataset fine-tuning and standard adapter training (freezing the Transformer parameters).
For context, we also compare with a method from recent work, \ti{dynamic sampling}~\citep{gottumukkala2020dynamic}, by sampling from each dataset in proportion to the difference between the current validation accuracy (EM+F1) on that dataset and the best accuracy from single-dataset training.
We train all models by sampling up to 75k training and 1k development examples from each dataset, following~\citet{fisch2019mrqa}. 
More details are in Appendix~\ref{appendix:training_details}.

Table~\ref{tab:in_domain} shows that {\multiadapter} scores higher than both single- and multi-dataset baselines.
Both phases of {\multiadapter} training---joint optimization followed by separate adapter tuning---are important for getting high accuracy.
Jointly optimizing the underlying {\multiadapter} Transformer improves performance compared to single-dataset adapters, suggesting that joint training encodes some useful cross-dataset information in the Transformer model.
Adapter tuning is important because the multi-dataset model converges on different datasets at different times, making it hard to find a single checkpoint that maximizes performance on all datasets (see Appendix Figure~\ref{fig:training_curve} for the training curve).
Some of the improvements can also be attributed to the adapter architecture itself, which slightly outperforms fine-tuning in most datasets.
Dynamic sampling does not improve results, possibly because the datasets are already balanced in size.

\begin{table*}[t!]
\centering
\resizebox{0.95\linewidth}{!}{%
\begin{tabular}{l l | ccccccc }
\toprule
\tf{$K$} & \tf{Model} & \tf{BioASQ} & \tf{DROP} & \tf{DuoRC} & \tf{RACE} & \tf{RelEx} & \tf{TextbookQA} & \tf{Avg.}\\
\midrule
\multirow{4}{*}{16}
 & \unifiedqa & 59.7 & 45.2 & 30.2 & \tf{52.9} & 82.0 & 33.8 & 50.6 \\
 & \robertamulti & 64.7 & 52.2 & 63.2 & 47.5 & 87.3 & 57.9 & 62.1 \\
 & {\multiadapter} ({\preaverage}) & \tf{67.0} & \tf{52.8} & 65.7 & 47.5 & \tf{87.7} & \tf{59.8} & 63.4 \\
 & {\multiadapter} ({\postaverage}) & 66.8 & 52.6 & \tf{67.2} & 47.3 & 87.7 & 59.7 & \tf{63.5} \\
\midrule
\multirow{4}{*}{64}
 & \unifiedqa & 65.5 & 47.9 & 32.9 & \tf{54.3} & 83.4 & 35.6 & 53.3 \\
 & \robertamulti & 66.1 & 54.9 & 64.3 & 48.3 & 87.8 & \tf{60.7} & 63.7 \\
 & {\multiadapter} ({\preaverage}) & 69.4 & \tf{55.9} & 66.8 & 47.6 & 88.4 & 60.2 & 64.7 \\
 & {\multiadapter} ({\postaverage}) & \tf{70.9} & 55.2 & \tf{66.9} & 47.5 & \tf{88.5} & 60.0 & \tf{64.8} \\
\midrule
\multirow{4}{*}{256}
 & \unifiedqa & 71.7 & 49.8 & 36.5 & \tf{55.2} & 85.6 & 36.8 & 55.9 \\
 & \robertamulti & 73.9 & \tf{57.2} & 65.2 & 49.1 & 89.1 & \tf{62.9} & 66.2 \\
 & {\multiadapter} ({\preaverage}) & 73.9 & 56.9 & 65.3 & 48.7 & 88.9 & 61.5 & 65.9 \\
 & {\multiadapter} ({\postaverage}) & \tf{75.1} & 56.9 & \tf{66.9} & 48.9 & \tf{90.2} & 61.8 & \tf{66.6} \\
\bottomrule
\end{tabular}
}
\caption{
\label{tab:transfer}
Transfer learning to MRQA out-of-domain datasets with $K$ training examples (F1, averaged over three random seeds). $^\dagger$: RACE is part of the UnifiedQA training data. 
pre avg.: average the adapter parameters at initialization. post avg.: fine-tune the adapters and jointly and average the parameters at the end. See Section~\ref{sec:transfer}.
}
\end{table*}

\subsection{Zero-shot Generalization}
\label{sec:zero_shot}

Table~\ref{tab:zero_shot} shows the results of applying this model to an unseen dataset (zero-shot).
We compare a simple method for using {\multiadapter}---averaging the parameters of the different adapters---with the multi-dataset model from Section~\ref{sec:in_distribution}, averaging the parameters of single-dataset adapters, and the pre-trained UnifiedQA-base~\citep{khashabi2020unifiedqa}.\footnote{
UnifiedQA was trained on different datasets with a different architecture, but represents an alternative off-the-shelf model for QA transfer learning.
We compare to UnifiedQA-base because the encoder has approximately the same number of parameters as RoBERTa-base.
}
We compare {\multiadapter} with and without the second phase of separate adapter-tuning.

Surprisingly, averaging the parameters of the different {\multiadapter} adapters results in a good model, generalizing better on average compared to both multi-dataset models.
The second phase of adapter-tuning improves these results.
Parameter averaging performs poorly for single-dataset adapters, possibly because the separately-trained adapters are too different from each other to interpolate well.

Figure~\ref{fig:zero_shot_comparison} compares the zero-shot accuracy obtained by the different {\multiadapter} and single-dataset adapters.
The two sets of adapters show similar patterns, with some adapters generalizing better than others, depending on the target, but all of the {\multiadapter} adapters generalize better than the corresponding single-dataset adapters.
This performance gap is considerably bigger than the gap in in-domain performance (Table~\ref{tab:in_domain}), further illustrating the benefit of joint optimization.

\subsection{Transfer Learning}
\label{sec:transfer}
Finally, we compare two ways of using {\multiadapter} for transfer learning: %
either averaging the adapter parameters and then fine-tuning the resulting model
(\ti{\preaverage}), or first fine-tuning all of the adapters and then taking the weighted average (\ti{\postaverage}). 
In both cases, we also back-propagate through the Transformer parameters.
We reserve 400 examples from each target dataset to use as a test set (following~\citealp{ram2021few}) and sample training datasets of different sizes, using half of the sampled examples for training and the other half as validation data for early stopping and to set the mixture weights for averaging the adapter parameters.

The results are in Table~\ref{tab:transfer}.
On average, {\multiadapter} leads to higher accuracy compared to the baselines, with bigger improvements for the smaller sizes of datasets, showing that a collection of robust single-dataset experts is a good starting point for transfer learning.
The post-average method performs about the same as averaging at initialization in the lower-data settings, and better with $K=256$.
All models struggle to learn with only 16 examples, and on DuoRC, which has long contexts and distant supervision and might represent a more challenging target for few-shot learning.
We also experimented with single-dataset adapters and with a frozen Transformer, which perform worse; detailed results are  in Appendix~\ref{appendix:transfer_learning_results_details}.

\section{Conclusion}
{\multiadapter} combines the benefits of single- and multi-dataset training, resulting in better in-domain accuracy, generalization, and transfer performance than either multi-dataset models or single-dataset models, especially in low resource settings.
For future work we plan to explore explicit mixture-modeling approaches for better zero-shot prediction and transfer learning.

\newpage
\section*{Acknowledgements}
We thank the members of the Princeton NLP group and the anonymous reviewers for their valuable comments. This work is supported by a Graduate Fellowship at Princeton University.

\bibliography{ref}
\bibliographystyle{acl_natbib}

\appendix

\clearpage
\section{Task Details}
\label{appendix:task_details}

\subsection{Dataset Details}
\label{appendix:dataset_details}

\begin{table*}[ht!]
\centering
\resizebox{1.2\columnwidth}{!}{%
\begin{tabular}{l l r r r}
\toprule
\tf{Dataset} & \tf{Domain} & $\bm{|c|}$ & \tf{\# train} & \tf{\# dev}  \\
\midrule
SQuAD 1.1 & Wikipedia & 137 & 86,588 & 10,507 \\
NewsQA & News articles & 599 & 74,160 & 4,212 \\
TriviaQA & Web snippets & 784 & 61,688 & 7,785 \\
SearchQA & Web snippets & 749 & 117,384 & 16,980 \\
HotpotQA & Wikipedia & 232 & 72,928 & 5,904 \\
Natural Questions & Wikipedia & 153 & 104,071 & 12,836 \\
\midrule
BioASQ & Science articles & 248 & - & 1,504 \\
DROP & Wikipedia & 243 & - & 1,503 \\
DuoRC & Movie plots & 681 & - & 1,501 \\
RACE & Examinations & 349 & - & 674 \\
RelationExtraction & Wikipedia & 30 & - & 2,948 \\
TextbookQA & Textbook & 657 & - & 1,503 \\
\bottomrule
\end{tabular}
}
\caption{
\label{tab:datasets}
Information about the MRQA datasets, from Table 1 of~\citet{fisch2019mrqa}, including the domain of the context passage; the average length of the context $c$ in tokens; and the number of training and development examples.
We downsample the in-domain datasets to 75K training and 1K development examples, and create few-shot training datasets from the out-of-domain datasets by reserving 400 examples as a test set and drawing different-sized training sets from the remaining examples, following~\citet{ram2021few}.
}
\end{table*}

We use the pre-processed datasets from the MRQA 2019 shared task~\citep{fisch2019mrqa}.
Table~\ref{tab:datasets} provides some dataset statistics.

\subsection{Training Details}
\label{appendix:training_details}
Our models are implemented in PyTorch~\citep{paszke2019pytorch} using HuggingFace~\citep{wolf2020transformers} and the adapter-transformers library~\citep{pfeiffer2020adapterhub}.
For all in-domain experiments, we sample 75,000 training and 1,000 validation examples and train with a constant learning rate and a batch size of 8, taking checkpoints every 1024 steps and stopping if validation F1 fails to improve for 10 checkpoints up to a fixed maximum number of epochs (10 for single-dataset training and 3 epochs for multi-dataset training).
We use a constant learning rate of 1e-5 for Transformer parameters and 1e-4 for adapter parameters, following standard settings for RoBERTa and adapters respectively~\citep{liu2019roberta,houlsby2019parameter}, and use the AdamW optimizer~\citep{loshchilov2018decoupled} with the HuggingFace default parameters.
For the multi-dataset models, we construct mini-batches of size $B$ by repeating $B$ times: pick a dataset uniformly, and pick an example uniformly from that dataset.

We train all models on single 2080Ti GPUs with 11GB of memory each.
The multi-dataset models take around two days to train, the single-dataset models take less than 24 hours, and it takes about 2 hours to train one model sequentially on six transfer datasets for three values of $K$ and three seeds.

\paragraph{Distant supervision}
Some datasets provide the gold answer string but do not mark the gold answer span in the context. We train the model to maximize the marginal likelihood of the gold answer string, marginalizing over all occurrences in the context. The set of possible answer spans are annotated in the pre-processed MRQA datasets.

\paragraph{Long contexts}
For inputs that are longer than the maximum input window for RoBERTa (512 tokens), we use a sliding window to split in the input into multiple ``chunks'': every input begins with the full question and the $\mathtt{[cls]}$ and separator tokens, and we fill the rest of the input window with tokens from the context, sliding the window 128 characters with each stride.
At prediction time, we return the answer from the chunk with that has the highest predicted probability.

\paragraph{Negative examples}
\label{sec:negative_examples}
We follow~\citet{longpre2019exploration} and include ``negative examples'' during training.
If a context chunk does not contain the answer span, we include the example as a training instance and train the model to indicate that the example does not contain the answer by selecting the $\mathtt{[cls]}$ token as the most likely start and end span.
At prediction time, we discard ``no answer'' predictions and return the non-empty answer from the chunk with that has the highest predicted probability.
For UnifiedQA, we train the model to predict an empty string for contexts that don't contain the answer to string and at prediction time return the non-empty answer with the highest probability.

\subsection{Transfer learning details}
\label{appendix:transfer_learning_details}
For transfer learning, we take 1/2 of the K training examples for validation and train for 200 steps or until the validation loss fails to improve for 10 epochs, and we reduce the adapter learning rate to 1e-5.
The other hyper-parameters are the same as for in-domain learning.

\paragraph{Training UnifiedQA}
We download the pre-trained UnifiedQA-base model from HuggingFace and train it in the format described in~\citet{khashabi2020unifiedqa} and in the accompanying code release.~\footnote{https://github.com/allenai/unifiedqa} We lower-case the question and context strings and concatenate them with a special string ``\textbackslash n'', which represents the backslash character followed by the letter n; and train the model to generate the answer string by minimizing cross-entropy loss.
We use greedy decoding for prediction.
In our pilot experiments, the recommended optimizer (Adafactor with a learning rate of 1e-3) quickly over-fits, so we use the same optimizer, learning rate, and batch size as for RoBERTa.

\section{Detailed Results}
\label{appendix:detailed_results}

\subsection{In-distribution Details}
\label{appendix:in_distribution_details}

\begin{figure*}
    \centering
    \resizebox{0.8\linewidth}{!}{%
    \includegraphics{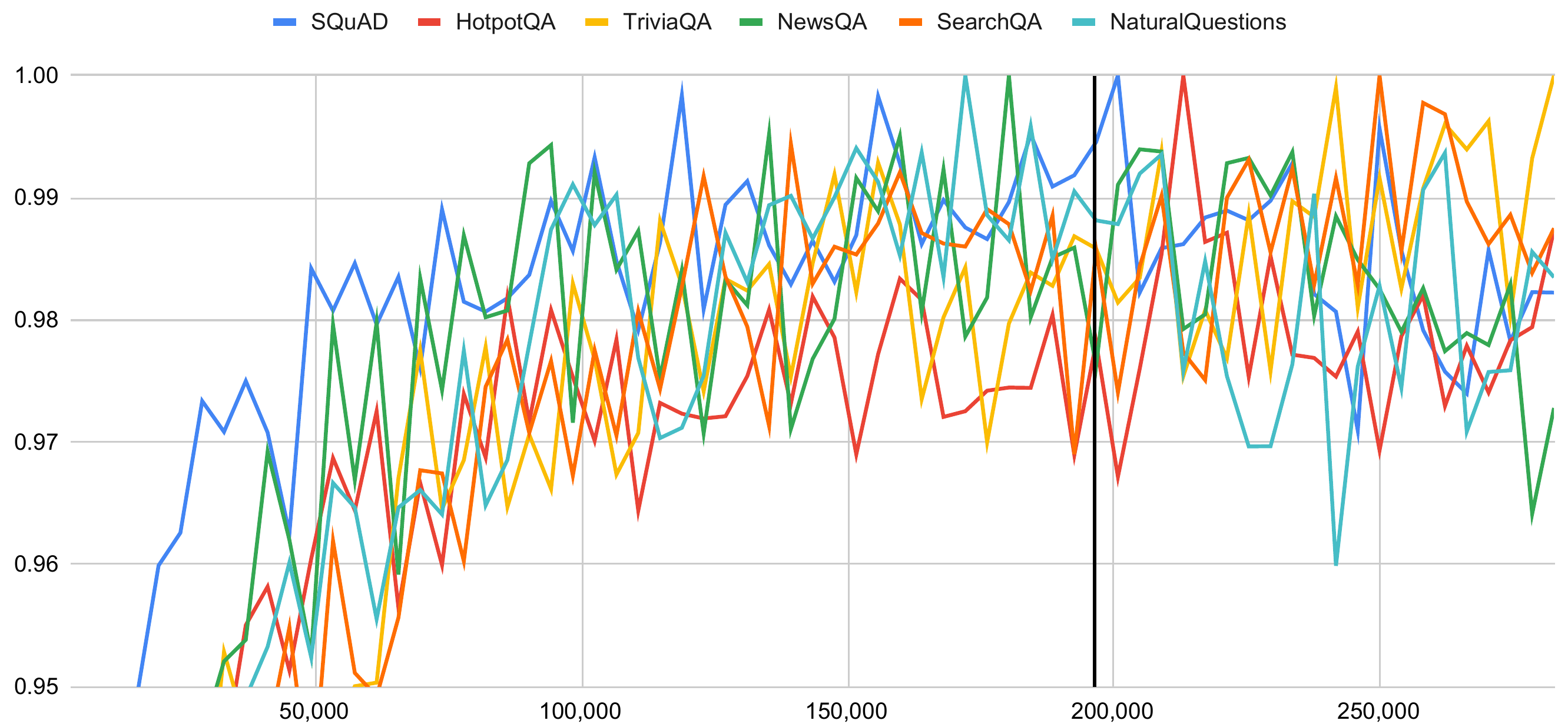}
    }
    \caption{
    Validation F1 from the joint-optimization phase of {\multiadapter} training, as a percentage of the maximum attained over the course of this run, plotted over number of optimization steps.
    The vertical black bar marks the checkpoint with the highest average F1, which is the checkpoint we select to keep, freezing the Transformer and continuing to tune the adapters separately (not shown).
    }
    \label{fig:training_curve}
\end{figure*}

Figure~\ref{fig:training_curve} shows the training curve for training {\multiadapter}, normalized by dividing each checkpoint score by the maximum validation accuracy obtained on that dataset during this run. The model reaches the maximum performance on the ``easy'' datasets early in training, which means that the model might over-fit to those datasets before converging on the more difficult datasets.
{\multiadapter} avoids this problem by tuning the adapter parameters separately after joint optimization.
Interestingly, adapter-tuning leads to improved performance on all datasets (Table~\ref{tab:in_domain}), even datasets on which joint-optimization appears to have already converged.

\subsection{Transfer Learning Details}
\label{appendix:transfer_learning_results_details}

\begin{table*}[t!]
\centering
\resizebox{1.0\linewidth}{!}{%
\begin{tabular}{l l | ccccccc }
\toprule
\tf{$K$} & \tf{Model} & \tf{BioASQ} & \tf{DROP} & \tf{DuoRC} & \tf{RACE} & \tf{RelEx} & \tf{TextbookQA} & \tf{Avg.}\\
\midrule
\multirow{7}{*}{16}
 & \unifiedqa & 59.7 & 45.2 & 30.2 & \tf{52.9}$^\dagger$ & 82.0 & 33.8 & 50.6 \\
 & \robertamulti & 64.7 & 52.2 & 63.2 & 47.5 & 87.3 & 57.9 & 62.1 \\
 & {\adapter} ({\preaverage}) & 63.0 & 37.7 & 58.8 & 35.8 & 84.1 & 48.8 & 54.7 \\
 & {\adapter} ({\postaverage}) & 62.1 & 40.9 & 58.8 & 30.4 & 84.0 & 48.8 & 54.2 \\
 & {\multiadapter} ({\preaverage}, freeze $\theta$) & 66.7 & 52.8 & 66.0 & 47.4 & \tf{87.7} & 60.2 & 63.5 \\
 & {\multiadapter} ({\postaverage}, freeze $\theta$) & 66.5 & 52.6 & 66.8 & 47.5 & \tf{87.7} & \tf{60.4} & \tf{63.6} \\
 & {\multiadapter} ({\preaverage}) & \tf{67.0} & \tf{52.8} & 65.7 & 47.5 & \tf{87.7} & 59.8 & 63.4 \\
 & {\multiadapter} ({\postaverage}) & 66.8 & 52.6 & \tf{67.2} & 47.3 & \tf{87.7} & 59.7 & 63.5 \\
\midrule
\multirow{7}{*}{64}
 & \unifiedqa & 65.5 & 47.9 & 32.9 & \tf{54.3}$^\dagger$ & 83.4 & 35.6 & 53.3 \\
 & \robertamulti & 66.1 & 54.9 & 64.3 & 48.3 & 87.8 & \tf{60.7} & 63.7 \\
 & {\adapter} ({\preaverage}) & 62.9 & 42.3 & 59.4 & 36.9 & 87.0 & 53.1 & 56.9 \\
 & {\adapter} ({\postaverage}) & 60.7 & 39.2 & 59.0 & 36.4 & 84.1 & 49.6 & 54.8 \\
 & {\multiadapter} ({\preaverage}, freeze $\theta$) & 66.9 & 53.3 & \tf{67.4} & 47.9 & 87.9 & 60.3 & 64.0 \\
 & {\multiadapter} ({\postaverage}, freeze $\theta$) & 65.4 & 52.2 & 66.8 & 47.9 & 87.8 & 59.9 & 63.3 \\
 & {\multiadapter} ({\preaverage}) & 69.4 & \tf{55.9} & 66.8 & 47.6 & 88.4 & 60.2 & 64.7 \\
 & {\multiadapter} ({\postaverage}) & \tf{70.9} & 55.2 & 66.9 & 47.5 & \tf{88.5} & 60.0 & \tf{64.8} \\
\midrule
\multirow{7}{*}{256}
 & \unifiedqa & 71.7 & 49.8 & 36.5 & \tf{55.2}$^\dagger$ & 85.6 & 36.8 & 55.9 \\
 & \robertamulti & 73.9 & \tf{57.2} & 65.2 & 49.1 & 89.1 & \tf{62.9} & 66.2 \\
 & {\adapter} ({\preaverage}) & 69.7 & 45.0 & 60.6 & 41.0 & 88.4 & 55.0 & 60.0 \\
 & {\adapter} ({\postaverage}) & 71.3 & 44.4 & 58.8 & 35.8 & 86.5 & 49.2 & 57.7 \\
 & {\multiadapter} ({\preaverage}, freeze $\theta$) & 66.8 & 53.5 & 66.9 & 48.0 & 87.9 & 60.2 & 63.9 \\
 & {\multiadapter} ({\postaverage}, freeze $\theta$) & 66.6 & 52.0 & \tf{67.0} & 47.8 & 87.8 & 59.9 & 63.5 \\
 & {\multiadapter} ({\preaverage}) & 73.9 & 56.9 & 65.3 & 48.7 & 88.9 & 61.5 & 65.9 \\
 & {\multiadapter} ({\postaverage}) & \tf{75.1} & 56.9 & 66.9 & 48.9 & \tf{90.2} & 61.8 & \tf{66.6} \\
\bottomrule
\end{tabular}
}
\caption{
\label{tab:transfer_details}
Transfer learning to MRQA out-of-domain datasets with $K$ training examples (F1, averaged over three random seeds), using the {\multiadapter} model with adapter-tuning. $^\dagger$: RACE is part of the UnifiedQA training data. pre avg.: we take the weighted average of adapters at initialization before fine-tuning; post avg.: we fine-tune each adapter jointly and average them at the end. \ti{Freeze $\theta$} refers to experiments where we freeze the Transformer parameters rather than tuning them along with the adapters.
}
\end{table*}

Table~\ref{tab:transfer_details} provides additional transfer learning results.
Single-dataset adapters transfer worse than {\multiadapter}, although performance improves considerably compared to zero-shot performance (Table~\ref{tab:zero_shot}).
We observe that the transfer process heavily down-weights some single-dataset adapters (like TriviaQA and SearchQA) that get high loss either before or after training, which might explain the performance improvement.
Freezing the Transformer parameters slighly improves results in the $K=16$ setting but leads to worse performance with more data.
The biggest drop is on BioASQ, possibly because it introduces new vocabulary and it is beneficial to update the token embeddings.

\end{document}